\documentclass[sigconf]{acmart}
\usepackage{booktabs} % For formal tables
\usepackage{subfigure}
\usepackage{bm}
\usepackage{xspace}
\usepackage{multirow}
\usepackage{footmisc}

\usepackage{amssymb}
\usepackage{array}
\usepackage{graphicx}
\usepackage{clrscode}
\usepackage{balance}
\usepackage{subfigure}
\usepackage{multirow}
\usepackage{multicol}
\usepackage{float}
\usepackage{color}
\usepackage{xcolor}
\usepackage{amsopn}
\usepackage{mathrsfs}
\usepackage{mathtools}
\usepackage{amsmath}
\usepackage{booktabs}
\usepackage{arydshln}
\usepackage{hyperref}
\usepackage{blkarray}
\usepackage{enumerate}
\usepackage{courier}
\usepackage{mathrsfs}
\usepackage{rotating}
\usepackage{bm}
\usepackage{subfigure}
\usepackage{array}
\usepackage{ragged2e}
\usepackage{amsthm}
\usepackage[ruled,linesnumbered]{algorithm2e}

\newcommand{\ie}{\emph{i.e.,}\xspace}

\newcommand{\eg}{\emph{e.g.,}\xspace}

\newcommand{\ignore}[1]{}

\AtBeginDocument{%
	\providecommand\BibTeX{{%
			\normalfont B\kern-0.5em{\scshape i\kern-0.25em b}\kern-0.8em\TeX}}}

\copyrightyear{2020}
\acmYear{2020}
\setcopyright{acmcopyright}
\acmConference[KDD '20] {26th ACM SIGKDD Conference on Knowledge Discovery and Data Mining}{August 23--27, 2020}{Virtual Event, USA}
\acmBooktitle{26th ACM SIGKDD Conference on Knowledge Discovery and Data Mining (KDD '20), August 23--27, 2020, Virtual Event, USA}
\acmPrice{15.00}
\acmDOI{10.1145/3394486.3403143}
\acmISBN{978-1-4503-7998-4/20/08}
\begin{document}
\fancyhead{}
%%
%% The "title" command has an optional parameter,
%% allowing the author to define a "short title" to be used in page headers.
\title{Improving Conversational Recommender Systems via\\ Knowledge Graph based Semantic Fusion}

\author{Kun Zhou}
\affiliation{%
  \institution{Peking University}}
\email{franciszhou@pku.edu.cn}

\author{Wayne Xin Zhao*$\dagger$}
\affiliation{%
  \institution{Gaoling School of Artificial Intelligence, Renmin University of China}}
%\additionalaffiliation{%
%\department[0]{Beijing Key Laboratory of Big Data Management and Analysis Methods.}}
\email{batmanfly@gmail.com}
\thanks{$^*$Corresponding author.}
\thanks{$^\dagger$Also with Beijing Key Laboratory of Big Data Management and Analysis Methods}

\author{Shuqing Bian}
\affiliation{
  \institution{School of Information, Renmin University of China}}
\email{bianshuqing@ruc.edu.cn}
%School of Information, 

\author{Yuanhang Zhou}
\affiliation{
  \institution{School of Information, Renmin University of China}}
\email{sdzyh002@gmail.com}

\author{Ji-Rong Wen$\dagger$}
\affiliation{
  \institution{Gaoling School of Artificial Intelligence, Renmin University of China}}
\email{jrwen@ruc.edu.cn}

\author{Jingsong Yu}
\affiliation{
  \institution{Peking University}}
\email{yjs@ss.pku.edu.cn}

\renewcommand{\shortauthors}{Zhou and Zhao, et al.}

\begin{abstract}
Conversational recommender systems (CRS) aim to recommend high-quality items to users through interactive conversations. Although several efforts have been made for CRS, two major issues still remain to be solved. 
First, the conversation data itself lacks of sufficient contextual information for accurately understanding users' preference.
Second, there is a semantic gap between natural language expression and item-level user preference. 
%utterance text is presented in natural language, while the user preference is reflected over the items or entities. There is a semantic gap 
%between need to enrich the contextual information and bridge the semantic gap 
%between the two kinds of data signals.

To address these issues, we incorporate both word-oriented and entity-oriented knowledge graphs~(KG) to enhance the data representations in CRSs, and adopt Mutual Information Maximization to align the word-level and entity-level semantic spaces.
%respectively. Furthermore, we propose to use Mutual Information Maximization to align the semantic space for the two components. 
Based on the aligned semantic representations, we further develop a KG-enhanced recommender component for making accurate recommendations, and a KG-enhanced dialog component that can generate informative keywords or entities in the response text.
Extensive experiments have demonstrated the effectiveness of our approach in yielding better performance on both recommendation and conversation tasks.
\end{abstract}




\maketitle

\section{Introduction}
Recently, conversational recommender system (CRS)~\cite{Li2018TowardsDC,Chen2019TowardsKR,Liao2019DeepCR,SunZ18,zhang2019building} has become an emerging research topic in seeking to provide high-quality recommendations through conversations with users. Different from traditional recommender systems, it emphasizes interactive clarification and explicit feedback in natural languages, and has a high impact on e-commerce.
%Due to its unique characteristics, CRS has been widely recognized to have a high impact on e-commerce.

In terms of methodology, CRS requires a seamless integration between a recommender component and a dialog component. 
On one hand, the dialog component clarifies user intents and replies to the previous utterance with suitable responses. %informative natural language expressions. 
On the other hand, the recommender component learns user preference and recommends high-quality items based on contextual utterances.
To develop an effective CRS, several solutions have been proposed to integrate the two components, including belief tracker over semi-structured user queries~\cite{SunZ18,zhang2019building} and switching decoder for component selection~\cite{Li2018TowardsDC,Liao2019DeepCR}.

Although these studies have improved the performance of CRS to some extent, two major issues still remain to be solved.
First, %it is difficult to accurately understand user preference based on the conversation alone, which usually consists of a few sentences. 
a conversation mainly consists of a few sentences, lack of sufficient contextual information for accurately understanding user preference. 
As shown in Table~\ref{example}, a user is looking for scary movies similar to ``\emph{Paranormal Activity (2007)}'', where her/his preference is simply described by two short sentences. 
In order to capture the user's intent, we need to fully utilize and model the contextual information. In this example, it is important to understand the underlying semantics of the word ``\emph{scary}'' and the movie ``\emph{Paranormal Activity (2007)}''.
Apparently, it is difficult to obtain such fact information solely based on the utterance text.
Second, utterances are represented in natural languages, while actual user preference is reflected over the items or entities (\eg actor and genre). There is a natural semantic gap between the two kinds of data signals.
We need an effective semantic fusion way to \emph{understand} or \emph{generate} the utterances. 
As shown in Table~\ref{example}, the system has presented both the recommended movie and the reason for recommendation.
Without bridging the semantic gap, it is infeasible to generate the text for explaining the recommendation,  \eg ``\emph{thriller movie with good plot}''.

\begin{table}
\caption{An illustrative example of a user-system conversation for movie recommendation.
The mentioned movies and important context words are marked in italic blue font and red font, respectively. }
\centering
\small
  \begin{tabular}{l|p{5.5cm}}
    \hline
    %\texttt{User} &Hello\\
   % \texttt{System} &Hi! How are you?\\
    \texttt{User} & I am looking for some movies\\
    \texttt{System} &What kinds of movie do you like? \\
    \hline
    \texttt{User} & Today I'm in a mood for something \textcolor{red}{scary}. Any similar movies like \textcolor{blue}{\emph{Paranormal Activity (2007)}}? \\
    \hline
    \texttt{System} &\textcolor{blue}{\emph{It (2017)}} might be good  for you. It is a classic \textcolor{red}{thriller} movie with \textcolor{red}{good plot}.\\
  %  \texttt{User} & I have seen that. Any other suggestions?  \\
    %\hline
    %\texttt{System} &I recommend \textcolor{blue}{\emph{The Conjuring 2 (2016)}}. It's a \textcolor{red}{shock horror film}. \\
    \hline
    \texttt{User} &Great! Thank you!\\
    \bottomrule
  \end{tabular}
  \label{case}
\label{example}
\end{table}

For enriching the conversation information, external knowledge graph~(KG) has been utilized in CRS~\cite{Chen2019TowardsKR}.  They mainly focus on incorporating item knowledge, while the word-level enrichment  (\eg the relation between ``\emph{scary}'' and ``\emph{thriller}'') has been somehow neglected. 
Furthermore, they have not considered the semantic gap between natural language and external knowledge. Therefore, the utilization of KG data is likely to be limited.
In essence, the problem originates from the fact that the dialog component and the recommender component correspond to two different semantic spaces, namely word-level and entity-level semantic spaces. 
Our idea is to incorporate two special KGs for enhancing data representations of both components, and fuse the two semantic spaces by associating the two KGs.

To this end, in this paper, we propose a novel conversational recommendation approach via KG based semantic fusion.
Specially, we incorporate a word-oriented KG (\ie ConceptNet~\cite{speer2016conceptnet}) and an item-oriented KG (\ie DBpedia~\cite{Bizer2009DBpediaA}). ConceptNet provides the relations between words, such as the synonyms, antonyms and co-occurrence words of a word; DBpedia provides the structured facts regarding the attributes of items.
We first apply graph neural networks to learn node embeddings over the two KGs separately, and then propose to apply the
Mutual Information Maximization~\cite{velikovi2018deep,Sun2019InfoGraphUA,Yeh2019QAInfomaxLR} method to bridge the semantic gap between the two KGs. The core idea is to force the representations of nodes in the two KGs to be close given the word-item co-occurrence  in the conversation.
In this way, we can unify the data representations in the two semantic spaces. 
Such a step is particularly useful to connect contextual words with items (including the mentioned entities) in conversations.
Based on the aligned semantic representations, we further develop a KG-enhanced recommender component for making accurate recommendations, and a KG-enhanced dialog component that can generate informative keywords or items in the response text.

To our knowledge, it is the first time that the integration of dialog and recommender systems has been addressed by using KG-enhanced semantic fusion. Our model utilizes two different KGs to enhance the semantics of words and items, respectively, and unifies their representation spaces.
Extensive experiments on a public CRS dataset have demonstrated the effectiveness of our approach in both recommendation and conversation tasks.

\section{Related Work}
Conversational recommendation system (CRS) contains two major modules, namely the recommender component and the dialog component. We first introduce the related work in the two aspects. 

Recommender systems aim to identify a subset of items that meet the user's interest from the item pool. Traditional methods are highly based on the historical user-item interaction~(\eg click and purchase)~\cite{survey,SVD++}. However, user-item interaction data is usually sparse.
To tackle the data sparsity problem, many techniques have been developed by utilizing the side information of items, such as review and taxonomy data~\cite{WangB11,BaoFZ14,huang-wsdm-2019}. As a comparison, CRS mainly focuses on the recommendation setting through conversation instead of historical interaction data.
Especially, knowledge graphs have been widely adopted to enhance the recommendation performance and explainability~\cite{huang-SIGIR-2018,Zhao-DI-2019,wang-aaai-2019}.  
%, while it may or may not use the user-item interaction data.

Conversation systems aim to generate proper responses given multi-turn contextual utterances. Existing works can be categorized into retrieval-based methods~\cite{Ji2014AnIR,zhou2016multi} and generation-based methods~\cite{Serban2015BuildingED,Li2015ADO}. The first category of approaches try to find the most reasonable response from a large repository of historical conversations~\cite{Ji2014AnIR,Wu2016SequentialMN}, and the generation-based methods utilize learnable models to produce the response text.
Based on the attentive seq2seq architecture~\cite{vinyals2015neural}, various extensions have been made to tackle the ``safe response'' problem and generate informative responses~\cite{Li2015ADO,Zhou2018CommonsenseKA,serban2017hierarchical,DBLP:conf/emnlp/ZhouZWLY19}.

Early conversational or interactive recommendation systems mainly utilized predefined actions to interact with users~\cite{Christakopoulou16}. Recently, several studies started to integrate the two components for understanding users' needs and recommend the right items through natural language conversation~\cite{SunZ18,ZhangCA0C18,Lei0MWHKC20}.
Overall, these methods emphasize the precise recommendation, while the conversation component is implemented by simple or heuristic solutions. 
Specially, a standard CRS dataset has been released in \cite{Li2018TowardsDC}, and a hierarchical RNN model was proposed for utterance generation. 
%. In this work, the authors also presented an utterance generation model with hierarchical RNN. However, they  relied on the training data to learn such a generation model.
Furthermore, follow-up studies~\cite{Chen2019TowardsKR,Liao2019DeepCR} incorporated external KG to improve the CRS, where their focus was to mainly enhance the item representations.

Based on previous studies, we design a novel conversational recommendation 
approach by incorporating and fusing word-level and entity-level knowledge graphs. Via KG fusion, our model is able to learn better data representations in both recommender and dialog components, which leads to better performance on item recommendation and utterance generation. 
%They have not considered the semantic fusion between natural language and external knowledge. While, our work incorporates both a word-oriented KG and an item-oriented KG, and fuses their semantic representations, which can make the representations mutually enhanced.
\section{Preliminaries}
Conversational recommendation systems (CRS) aim to recommend proper items to a user through a multi-turn conversation.
In the conversation, a chat agent analyzes and learns the preference of the user according to contextual conversation history. Then, it either generates appropriate recommendations or starts a new round of conversation for further clarification.
The process ends until the task succeeds or the user leaves.
In a CRS, there are two major components to develop, namely the recommender component and the dialog component.
The two components should be integrated seamlessly, and successful recommendation is considered as the final goal.

Formally, let $u$ denotes a user from user set $\mathcal{U}$, $i$ denotes an item from item set $\mathcal{I}$, and $w$ denotes a word from vocabulary $\mathcal{V}$.
A conversation (or a conversation history) $C$ consists of a list of utterances, denoted by $C=\{s_{t}\}_{t=1}^{n}$, in which each utterance $s_t$ is a conversation sentence at the $t$-th turn.
At the $t$-th turn, the recommender component selects a set of candidate items $\mathcal{I}_t$ from the entire item set $\mathcal{I}$ according to some strategy, while the dialog component needs to produce the next utterance $s_t$ to reply to previous utterances.
Note that $\mathcal{I}_t$ can be equal to $\emptyset$ when there is no need for recommendation. In such a case, the dialog component may raise a clarification question or generate a chit-chat response.
Given a $n$-turn conversation, the goal of a CRS is to generate the \emph{response utterance} to the user, including both the recommendation set $\mathcal{I}_{n+1}$
and the reply utterance $s_{n+1}$.

\section{Approach}
\label{CRS}
\begin{figure*}
\includegraphics[width=0.95\textwidth]{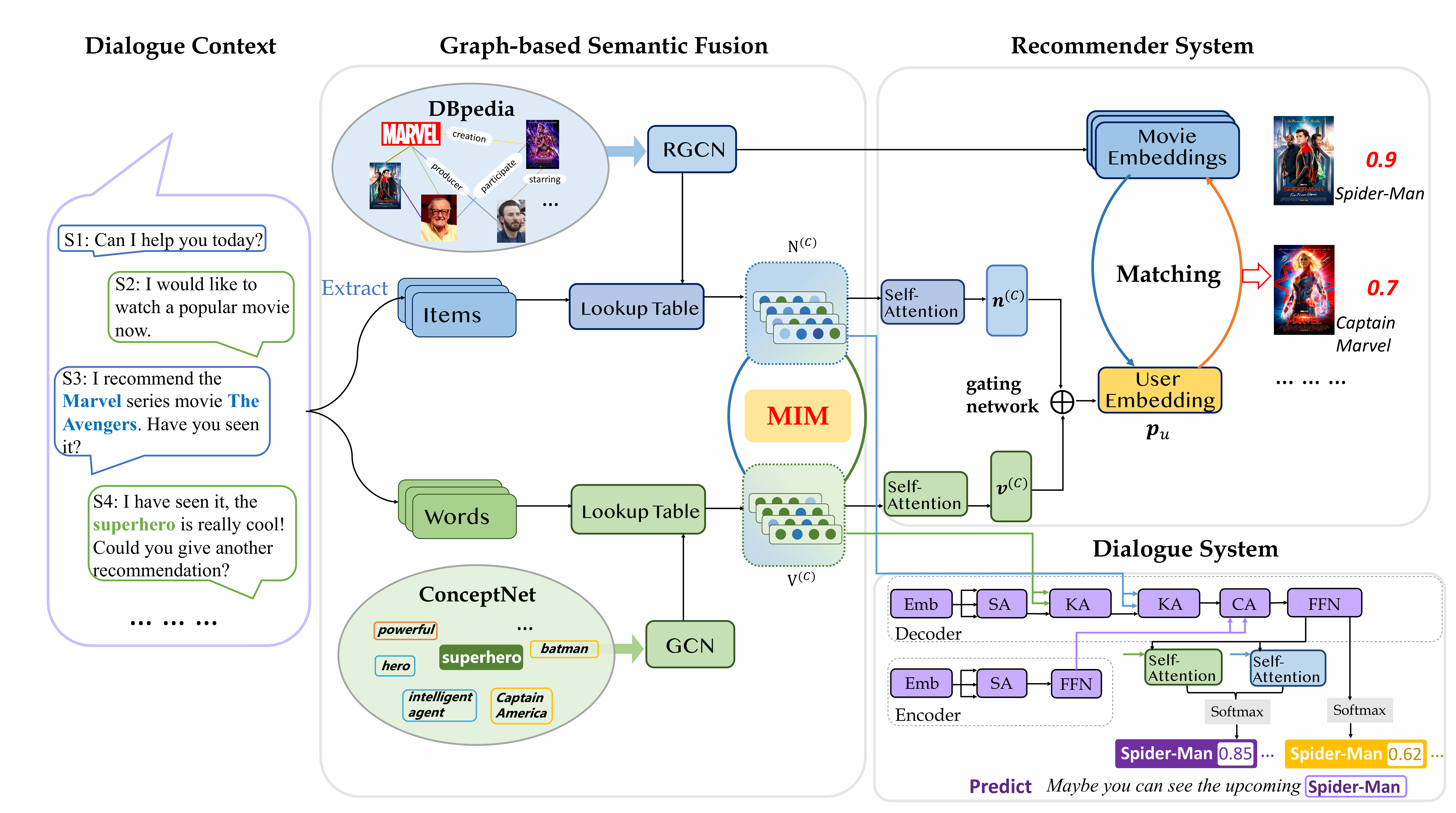}
\caption{The overview of our model with a movie recommendation scenario. Here,``SA'',   ``KA'',  and ``CA'' denotes \emph{self-attention}, \emph{KG-based attention} and \emph{context-based attention}, respectively.}
\label{approach}
\end{figure*}

In this section, we present the \underline{KG}-based \underline{S}emantic \underline{F}usion approach to the CRS task, named \emph{KGSF}. We first introduce how to encode both word-oriented and item-oriented KGs, and then fuse the semantics of the two KGs. Based on the fused KGs, we finally describe our solutions for both recommendation and conversation tasks.  
The overview illustration of the proposed model is presented in Fig.~\ref{approach}.

\subsection{Encoding External Knowledge Graphs}
As shown in Table~\ref{example}, it is difficult to fully understand user preference and generate a suitable response.  
We identify the two basic semantic units in dialog and recommender systems, namely \emph{word} and \emph{item}, respectively.
Here, we utilize two separate KGs to enhance the representations of the basic semantic units on both sides.

\subsubsection{Encoding Word-oriented KG}
\label{word-graph}
We adopt the widely used ConceptNet~\cite{speer2016conceptnet} as the word-oriented KG. It stores a semantic fact as a triple $\langle w_1, r, w_2\rangle$, where $w_1, w_2 \in \mathcal{V}$ are words and $r$ is a word relation.
Not all the words in ConceptNet are useful for our task. Hence, we only consider words that appear in our corpus, and extract their related triples from ConceptNet. We also remove words with very few triples related to the words in our corpus.

To encode the word-oriented KG, we adopt graph convolutional neural network~\cite{KipfW16,EdwardsX16}~(GCN) for capturing the semantic relations between word nodes.
At each update, GCN receives the information from the one-hop neighborhood in the graph and performs the aggregation operation as:
\begin{eqnarray}
\label{gcn}
\mathbf{V}^{(l)}=\text{ReLU}(\mathbf{D}^{-\frac{1}{2}}\mathbf{A}\mathbf{D}^{-\frac{1}{2}}\mathbf{V}^{(l-1)}\mathbf{W}^{(l)})
\end{eqnarray}
where $\mathbf{V}^{(l)}\in \mathbb{R}^{V \times d_W}$ are the representations of nodes and $\mathbf{W}^{(l)}$ is a learnable matrix at the $l$-th layer, $\mathbf{A}$ is the adjacency matrix of the graph and $\mathbf{D}$ is a diagonal degree matrix with entries $\mathbf{D}[i,i]=\sum_{j}\mathbf{A}[i,j]$. By stacking multiple convolutions, node information can be propagated along with the graph structure. When the algorithm ends, we can obtain a $d_W$-dimensional representation $\bm{n}_w$ for a word $w$.
Here, we do not incorporate the relation information, because the number of relations is large and many relations are not directly useful to the recommendation task. 

\subsubsection{Encoding Item-oriented KG}
\label{entity-graph}
Another kind of semantic units to consider are items and their related entities. Following~\cite{Chen2019TowardsKR}, we utilize  DBpedia~\cite{Bizer2009DBpediaA} as item-oriented KG. Similar to ConceptNet, a triple in DBpedia is denoted by $\langle e_1,r,e_2\rangle$, where $e_1, e_2 \in \mathcal{E}$ are items or entities from the entity set $\mathcal{E}$ and $r$ is entity relation from the relation set $\mathcal{R}$.
To extract the entity subgraph, we collect all the entities appearing in our corpus by following the approach from~\cite{Chen2019TowardsKR}. 
Starting from these items and entities as seeds, we extract their one-hop triples on the DBpedia graph.

For item-oriented KG, relation semantics are important to consider. Different from ConceptNet, we utilize R-GCN~\cite{Schlichtkrull_2018} to learn item representations on the extracted subgraph.
Formally, the representation of node $e$ at $(l+1)$-th layer is calculated as:
\begin{equation}
\label{r-gcn}
\bm{n}_{e}^{(l+1)}=\sigma\big(\sum_{r \in \mathcal{R}}\sum_{e' \in \mathcal{E}_{e}^{r}}\frac{1}{Z_{e,r}}\mathbf{W}_{r}^{(l)}\bm{n}_{e'}^{(l)}+\mathbf{W}^{(l)}\bm{n}_{e}^{(l)}\big)
\end{equation}
where $\bm{n}_{e}^{(l)} \in \mathbb{R}^{d_E}$ is the node representation of $e$ at the $l$-th layer, $\mathcal{E}_{e}^{r}$ denotes the set of neighboring nodes for $e$ under the relation $r$, $\mathbf{W}_{r}^{(l)}$ is a learnable relation-specific transformation matrix for the embeddings from neighboring nodes with relation $r$, $\mathbf{W}^{(l)}$ is a learnable matrix for transforming the representations of nodes at the $l$-th layer and  $Z_{e,r}$ is a normalization factor.

\subsection{KG Fusion via Mutual Information Maximization}
\label{MIM}
Above, we obtain the node representations for the word-oriented KG and item-oriented KG, denoted by two embedding matrices $\mathbf{V}$ ($\bm{v}_w$ for word $w$) and $\mathbf{N}$ ($\bm{n}_e$ for item $e$), respectively.

In order to bridge the semantic gap between words and items,
we propose to use Mutual Information Maximization technique~\cite{velikovi2018deep,Sun2019InfoGraphUA,Yeh2019QAInfomaxLR}, called \emph{MIM}.
MIM has been used to mutually improve the data representations of two coupled signals~(\eg input and output).
Its core idea is based on the concept of Mutual Information~(MI).
Given two variables $X$ and $Y$, their MI is defined as:
\begin{eqnarray}
MI(X,Y)=D_{KL}(p(X,Y)||p(X)p(Y)),
\end{eqnarray}
where $D_{KL}$ is the Kullback-Leibler~(KL) divergence between the joint distribution $p(X,Y)$ and the product of marginals $p(X)p(Y)$. Usually, MI is difficult to compute.  MIM tries to maximize it instead of seeking the precise value via the following formula:
\begin{equation}\label{eq-infomax}
MI(X,Y)\ge \mathbb{E}_{P}[g(x,y)]- \mathbb{E}_{N}[g(x',y')],
\end{equation}
where $ \mathbb{E}_{P}$ and $ \mathbb{E}_{P}$ denote the expectation over positive and negative samples respectively, and $g(\cdot)$ is the binary classification function that outputs a real number which can be modeled by a neural network.

In our setting, we have two kinds of semantic units (namely words and entities), and would like to align their semantic representation spaces.
Given a conversation, we first collect words (non-stopwords) and entities (including items) from the utterance text.
For an entity-word pair $\langle e, w \rangle$ that co-occur in a conversation, we pull their representations close through a transformation matrix:
\begin{equation}\label{eq-gf}
g(e, w) = \sigma( \bm{n}_e^{\top} \cdot \mathbf{T} \cdot \bm{v}_w),
\end{equation}
where $ \bm{n}_e$ and $\bm{v}_w$ are the learned node representations for entity $e$ and word $w$ via KGs, respectively, $\mathbf{T} \in \mathbb{R}^{d_E \times d_W}$ is the transformation matrix that aligns two semantic spaces, and $\sigma(\cdot)$ is the sigmoid function.
To apply the MIM method, we can consider all the word-entity pairs co-occurring in a conversation as \emph{positive}, while random word-entity pairs are considered as \emph{negative}.
By integrating Eq.~\ref{eq-gf} into Eq.~\ref{eq-infomax}, we can derive the objective loss over all the conversations and minimize the loss with an optimization algorithm.

However, a conversation usually contains a number of contextual words, and it is time-consuming to enumerate all the word-entity pairs.
Besides, some of these words are noisy, which is likely to affect the final performance.
Here, we add a super token $\tilde{w}$ for a conversation, assuming that it is able to represent the overall semantics of the contextual words. Instead of considering all the word-entity pairs, we only model the relation between each entity and the super token $\tilde{w}$ using the $g(\cdot)$ function.
We utilize self-attention mechanism for learning the representation of $\tilde{w}$:
\begin{eqnarray}\label{eq-SA}
\bm{v}_{\tilde{w}} &=&  \mathbf{V}^{(C)} \cdot \bm{\alpha},\\\nonumber
\bm{\alpha} &=& \text{softmax}( \bm{b}^\top \cdot \text{tanh}(\mathbf{W}_{\alpha}  \mathbf{V}^{(C)})),
\end{eqnarray}
where $ \mathbf{V}^{(C)}$ is the matrix consisting of the embeddings of all the contextual words in a conversation $C$, $\bm{\alpha}$ is an attention weight vector reflecting the importance of each word, and $\mathbf{W}_{\alpha} $ and $\bm{b}$ are parameter matrix and vector to learn. Using such a super token, we can significantly improve efficiency and identify more important semantic information from the entire conversation.

In order to effectively align the semantic space of the two KGs, we adopt the MIM loss for pre-training the parameters of the GNN models in Section~\ref{word-graph} and \ref{entity-graph}, which forces the two semantic spaces to be close at the beginning. During the fine-tuning stage, we treat the MIM loss as a regularization constraint for GNN to prevent overfitting.

With the representations of the fused KGs, we next describe how to make recommendations and generate utterances in Section 4.3 and 4.4, respectively. 

\subsection{KG-enhanced Recommender Module}
Given the learned word and item representations, we study how to generate a set of items for recommendation in CRS.

A key point for recommendation is to learn a good representation of user preference.
Different from traditional recommender systems, following~\cite{Li2018TowardsDC,Chen2019TowardsKR}, we assume no previous interaction records are available.
We can only utilize the conversation data to infer user preference.

First, we collect all the words that appear in a conversation $c$.
By using a simple lookup operation, we can obtain the word or item embeddings learned through the graph neural networks in Section~\ref{MIM}. We concatenate the word embeddings into a matrix $\mathbf{V}^{(C)}$. Similarly, we can derive an item embedding matrix $\mathbf{N}^{(C)}$ by combining the embeddings of items.

Next, we apply the similar self-attentive mechanism in Eq.~\ref{eq-SA} to learn a single word vector $\bm{v}^{(C)}$ for $\mathbf{V}^{(C)}$ and a single item vector $\bm{n}^{(C)}$  for  $\mathbf{N}^{(C)}$. In order to combine the two parts of information, we apply the gate mechanism to derive the preference representation $\bm{p}_u$ of the user $u$:
\begin{eqnarray}
\label{gate}
\bm{p}_u &=& \beta \cdot \bm{v}^{(C)}  + (1-\beta) \cdot \bm{n}^{(C)},\\\nonumber
\beta &=&\sigma(\mathbf{W}_{\text{gate}}[\bm{v}^{(C)}; \bm{n}^{(C)} ]),\nonumber
\end{eqnarray}

Given the learned user preference, we can compute the probability that recommends an item $i$ from the item set to a user $u$:
\begin{equation}
\label{eq-rec}
\text{Pr}_{rec}(i) = \text{softmax}(\bm{p}_u^\top \cdot \bm{n}_i ),
\end{equation}
where $\bm{n}_i$ is the learned item embedding for item $i$. We can utilize Eq.~\ref{eq-rec} to rank all the items and generate a recommendation set to a user. 
To learn the parameters, we set a cross-entropy loss as:
\begin{eqnarray}
\label{CE-rec}
L_{rec}=-\sum_{j=1}^{N} \sum_{i=1}^{M} &&\big[-(1-y_{ij})\cdot\log\big(1-\text{Pr}^{(j)}_{rec}(i))\big) \\ \nonumber
&&+y_{ij}\cdot\log\big(\text{Pr}^{(j)}_{rec}(i)\big) 
\big]+\lambda*L_{MIM},
\end{eqnarray}
where $j$ is the index of a conversation, $i$ is the index of an item, $L_{MIM}$ is the Mutual Information Maximization loss, and $\lambda$ is a weighted parameter. Here, we loop the entire collection and compute the cross-entropy loss. We only present the case that a conversation has a ground-truth recommendation. While, it is straightforward to extend the above loss to the case with multiple ground-truth items. 

\subsection{KG-enhanced Response Generation Module}
Here, we study how to generate a reply utterance in CRS.
We adopt Transformer~\cite{vaswani2017attention} to develop the encoder-decoder framework.
Our encoder follows a standard Transformer architecture. We mainly introduce the KG-enhanced decoder.

To better generate responses at decoding, we incorporate KG-enhanced representations of context words and items. After the self-attention sub-layer, we conduct two KG-based attention layerss to fuse the information from the two KGs:
\begin{eqnarray}
\mathbf{A}^{n}_{0}&=&\text{MHA}(\mathbf{R}^{n-1},\mathbf{R}^{n-1},\mathbf{R}^{n-1}), \\
\mathbf{A}^{n}_{1}&=&\text{MHA}(\mathbf{A}^{n}_{0},\mathbf{V}^{(C)},\mathbf{V}^{(C)}), \label{eq-W}\\
\mathbf{A}^{n}_{2}&=&\text{MHA}(\mathbf{A}^{n}_{1},\mathbf{N}^{(C)},\mathbf{N}^{(C)}), \label{eq-C}\\
\mathbf{A}^{n}_{3}&=&\text{MHA}(\mathbf{A}^{n}_{2},\mathbf{X},\mathbf{X}), \label{eq-en}\\
\mathbf{R}^{n}&=&\text{FFN}(\mathbf{A}^{n}_{3}),\label{eq-T}
\end{eqnarray}
where $\text{MHA}(\mathbf{Q}, \mathbf{K}, \mathbf{V})$ defines the multi-head attention function~\cite{vaswani2017attention} that takes a query matrix $\mathbf{Q}$, a key matrix $\mathbf{K}$, and a value matrix $\mathbf{V}$ as input:
\begin{eqnarray}
\text{MHA}(\mathbf{Q}, \mathbf{K}, \mathbf{V})=\text{Concat}(\text {head}_{1}, \ldots, \text {head}_{\mathrm{h}}) \mathbf{W}^{O},\\
\text{head}_{i}=\text {Attention}(\mathbf{Q} \mathbf{W}_{i}^{Q}, \mathbf{K} \mathbf{W}_{i}^{K}, \mathbf{V} \mathbf{W}_{i}^{V}),
\end{eqnarray}
and $\text{FFN}(\bm{x})$ defines a fully connected feed-forward network, which consists of a linear transformation with a ReLU activation layer:
\begin{eqnarray}
\text{FFN}(\bm{x})=\max (0, \bm{x} \mathbf{W}_{1}+b_{1}) \mathbf{W}_{2}+b_{2}.
\end{eqnarray}
Above, $\mathbf{X}$ is the embedding matrix output by the encoder, $\mathbf{V}^{(C)}$ and $\mathbf{N}^{(C)}$ are KG-enhanced representation matrices for words and items in a conversation $c$, respectively. And, $\mathbf{A}^{n}_{0}$, $\mathbf{A}^{n}_{1}$, $\mathbf{A}^{n}_{2}$ and $\mathbf{A}^{n}_{3}$ are the representations after self-attention, cross-attention with embeddings from ConceptNet, cross-attention with embeddings from DBpedia  and cross-attention with encoder output, respectively. Finally, $\mathbf{R}^{n}$ is the embedding matrix from the decoder at $n$-th layer. 

Compared with a standard Transformer decoder, we have two additional steps in Eq.~\ref{eq-W} and \ref{eq-C}.
The idea can be described using a transformation chain:
\emph{generated words} $\xrightarrow{Eq.~\ref{eq-W}}$ \emph{word-oriented KG} $\xrightarrow{Eq.~\ref{eq-C}}$ \emph{item-oriented KG} $\xrightarrow{Eq.~\ref{eq-en}}$ \emph{context words}.
Following such a chain, our decoder is able to gradually inject useful knowledge information from the two KGs in a sequential manner.
The rationality for Eq.~\ref{eq-C} lies in the fact that we have fused the two KGs as in Section~\ref{MIM}.

Different from chit-chat models, the generated reply is expected to contain the recommended items, related entities and descriptive keywords. We further adopt the copy mechanism to enhance the generation of such tokens.
Formally, given the predicted subsequence $y_1, \cdots, y_{i-1}$, the probability of generating $y_i$ as the next token is given as:
\begin{eqnarray}
\text{Pr}(y_i | y_1, \cdots, y_{i-1} ) = \text{Pr}_{1}(y_i | \mathbf{R}_i ) + \text{Pr}_{2}(y_i | \mathbf{R}_i, \mathcal{G}_1, \mathcal{G}_2),
\label{gen-copy}
\end{eqnarray}
where $\text{Pr}_{1}(\cdot)$ is the generative probability implemented as a softmax function over the vocabulary by taking the decoder output $\mathbf{R}_i$ (Eq.~\ref{eq-T}) as input,  $\text{Pr}_{2}(\cdot)$ is the copy probability implemented by following a standard copy mechanism~\cite{DBLP:conf/acl/GuLLL16} over the nodes of the two KGs, and $\mathcal{G}_1, \mathcal{G}_2$ denote the two KGs we have used. 
To learn the response generation module, we set the cross-entropy loss as:
\begin{eqnarray}
L_{gen}=-\frac{1}{N}\sum_{t=1}^{N}\log\big(\text{Pr}(s_t | s_1, \cdots, s_{t-1}))\big),
\label{CE-gene}
\end{eqnarray}
where $N$ is the number of turns in a conversation $C$. We compute this loss for each utterance $s_t$ from $C$.

\subsection{Parameter Learning}

\begin{algorithm}[t]
\small
	\caption{The training algorithm for the KGFS model.}
	\label{algorithm}
	\LinesNumbered
	\KwIn{
		The conversation recommendation dataset $\mathcal{D}$, item-oriented KG $\mathcal{G}_{1}$, and word-oriented KG $\mathcal{G}_{2}$
	}
	\KwOut{Model parameters $\Theta^{g}$, $\Theta^{r}$ and $\Theta^{d}$.}
	Randomly initialize $\Theta^{g}$, $\Theta^{r}$ and $\Theta^{d}$.\\
	Pre-train the $\Theta^{g}$ by minimizing the MIM loss in~\ref{MIM}.\\
	\For{$t = 1 \to |\mathcal{D}|$}{	
		Acquire items' and words' representations from $\mathcal{G}_{1}$ and $\mathcal{G}_{2}$ by Eq.~\ref{gcn} and Eq.~\ref{r-gcn}, respectively.\\
		Acquire $\bm{v}^{(C)}$ and $\bm{v}^{(C)}$ by self-attention using Eq.~\ref{eq-SA}.\\
		Acquire $\bm{p}_{u}$ by gate mechanism using Eq.~\ref{gate}.\\
		Compute $\text{Pr}_{rec}(i)$ using Eq.~\ref{eq-rec}.\\
		Perform GD on Eq.~\ref{CE-rec} \emph{w.r.t.} $\Theta^{g}$ and $\Theta^{r}$.\\
	}
	\For{$i = 1 \to |\mathcal{D}|$}{
		Acquire items' and words' representations from $\mathcal{G}_{1}$ and $\mathcal{G}_{2}$ by Eq.~\ref{gcn} and Eq.~\ref{r-gcn}, respectively.\\
		Acquire $\mathbf{R}^{n}$ by KG-enhanced Transformer using Eq.~\ref{eq-T}.\\
		Compute $\text{Pr}(y|y_{1},\cdots,y_{i-1})$ using Eq.~\ref{gen-copy}.\\
		Perform GD on Eq.~\ref{CE-gene} \emph{w.r.t.} $\Theta^{d}$.\\
	}
	\Return $\Theta^{g}$, $\Theta^{r}$ and $\Theta^{d}$.
\end{algorithm}

Our parameters to learn are organized by three groups, namely the KG module, recommender module and conversation module, denoted by $\Theta^{g}$, $\Theta^{r}$ and $\Theta^{d}$ respectively. Algorithm~\ref{algorithm} presents the training algorithm for our KGSF model. The three components share parameters and affect each other. 

To train the joint model, we pre-train the knowledge graph module $\Theta^{g}$ using Mutual Information Maximization loss. Next, we optimize the parameters in $\Theta^{r}$ and $\Theta^{g}$ . At each iteration, we first acquire words' and items' representations from the KG module. Then, we perform the self-attention and gate mechanism to derive user representations. Finally, we compute a cross-entropy loss by Eq.~\ref{CE-rec} with the MIM regularization, and perform gradient descent to update parameters $\Theta^{r}$ and $\Theta^{g}$.

When the loss of the recommender component converges, we optimize the parameters in $\Theta^{d}$. At each iteration, we first obtain words' and items' representations from the knowledge graph module. Then, we utilize KG-enhanced Transformer to derive the contextual representations. Finally, we compute the cross-entropy loss by Eq.~\ref{CE-gene}, and perform gradient descent to update parameters in $\Theta^{d}$.

\section{Experiment}
In this section, we first set up the experiments, and then report the results and analysis. 
\subsection{Experiment Setup}
\begin{table}
\caption{Results on recommendation task. Numbers marked with * indicate that the improvement is statistically significant compared with the best baseline (t-test with p-value $< 0.05$). }\label{rec-result}
\small
\centering
\begin{tabular}{lcccccc}
    Test& \multicolumn{3}{c}{All data} & \multicolumn{3}{c}{Cold start} \\
    \hline
     Models &R@1 &R@10 &R@50 &R@1 &R@10 &R@50 \\
    \hline
    \texttt{Popularity} &0.012 &0.061 &0.179 &0.020 &0.097 &0.239\\
    \texttt{TextCNN} & 0.013 &0.068 &0.191 &0.011& 0.081& 0.239 \\
    \texttt{ReDial} & 0.024& 0.140& 0.320 & 0.021 &0.075 &0.201\\
    \texttt{KBRD} & 0.031& 0.150& 0.336 &0.026 &0.085 &0.242\\
    \hline
    \textbf{\texttt{KGSF}} & \textbf{0.039}*& \textbf{0.183}*& \textbf{0.378}* &\textbf{0.039}* &\textbf{0.174}* &\textbf{0.370}*\\
    \hline
    %\texttt{KGSF w/o MIM-ft} & 0.036 & 0.179 &0.366 &0.037 &0.166 &0.370\\
    \texttt{ -\emph{MIM}} & 0.037 & 0.175 &0.356 &0.037 &0.158 &0.331\\
    \texttt{ -\emph{DB}} & 0.027 & 0.121 &0.256 &0.030 &0.168 &0.346\\
    \hline
  \end{tabular}
\end{table}

\subsubsection{Dataset}
We evaluate our model on the \emph{REcommendations through DIALog (REDIAL)} dataset, which is a conversational recommendation dataset released by \cite{Li2018TowardsDC}. This dataset was constructed through Amazon Mechanical Turk (AMT). Following a set of comprehensive instructions, the AMT workers generated dialogs for recommendation on movies in a seeker-recommender pair. It contains 10,006 conversations consisting of 182,150 utterances related to 51,699 movies. This dataset is split into training, validation and test sets using a ratio of 8:1:1. 
For each conversation, we start from the first sentence one by one to generate  reply utterances or recommendations by our model.

\subsubsection{Baselines}
In CRS, we consider two major tasks for evaluation, namely recommendation and conversation. 

$\bullet$ \emph{Popularity}: It ranks the items according to historical recommendation frequencies in the corpus.

$\bullet$ \emph{TextCNN}~\cite{Kim14}: It adopts a CNN-based model to extract textual features from contextual utterances as user embedding.

$\bullet$ \emph{Transformer}~\cite{vaswani2017attention}: It applies a Transformer-based encoder-decoder framework to generate proper responses without information from recommender module.

$\bullet$ \emph{REDIAL}~\cite{Li2018TowardsDC}: This model has been proposed in the same paper with our dataset~\cite{Li2018TowardsDC}. It basically consists of a dialog generation module based on
HRED~\cite{serban2017hierarchical}, a recommender module based on auto-encoder~\cite{he2017distributedrepresentation} and a sentiment analysis module.

$\bullet$  \emph{KBRD}~\cite{Chen2019TowardsKR}: This model utilizes DBpedia to enhance the semantics of contextual items or entities. The dialog generation module is based on the Transformer architecture, in which KG information serves as word bias for generation.

Among these baselines, \emph{Popularity} and \emph{TextCNN}~\cite{Kim14} are recommendation methods, and \emph{Transformer}~\cite{vaswani2017attention} is the state-of-the-art text generation method. We do not include other recommendation models, since there are no historical user-item interaction records except the text of a single conversation.
Besides,  \emph{REDIAL}~\cite{Li2018TowardsDC} and \emph{KBRD}~\cite{Chen2019TowardsKR} are conversation recommendation methods. We name our proposed model as \emph{KGSF}.

\subsubsection{Evaluation Metrics}
In our experiments, we adopt different metrics to evaluate the two tasks.
For recommendation task, following~\cite{Chen2019TowardsKR}, we adopt Recall@$k$ ($k=1, 10, 50$) for evaluation.
Besides the standard setting, we consider a specific scenario in recommender systems, namely \emph{cold start}.

In CRS, this problem can be alleviated to some extent, since we have conversation contexts. 
In order to simulate the cold-start scenario in CRS, we only consider the test cases without any mentioned items in context. This experiment aims to examine whether our fusion strategy is useful to learn user preference from word-based utterances.
For the conversation task, the evaluation consists of automatic evaluation and human evaluation. Following~\cite{Chen2019TowardsKR}, we use Distinct $n$-gram ($n=2,3,4$) to measure the diversity at sentence level. For CRS, it is particularly important that the dialog system is able to generate informative replies related to items or entities. Hence, we introduce a new metric that calculates the ratio of items in the generated utterances. Different from traditional conversation tasks, we do not need to generate response resembling the ground-truth utterance. Instead, the final goal is to successfully make the recommendations. 
For this reason, we adopt human evaluation (on a random selection of 100 multi-turn dialogs from the test set) instead of using BLEU metrics.
We invite three annotators to score the generated candidates in two aspects, namely \emph{Fluency} and \emph{Informativeness}.
The range of score is 0 to 2.
The final performance is calculated using the average scores of the three annotators.

\subsubsection{Implementation Details} 
%By integrating the three components in Section~\ref{CRS}, we can obtain the complete KGSF model for  CRS. 
%We can follow Algorithm 1 for learning the parameters of our approach. 
We follow the procedure in Algorithm~\ref{algorithm} to implement our approach with Pytorch~\footnote{https://pytorch.org/}. The dimensionality of embeddings (including hidden vectors) is set to 300 and 128, respectively, for conversation and recommender modules. We initialize word embeddings via word2vec\footnote{https://radimrehurek.com/gensim/models/word2vec.html}. In the KG module, we set the layer number to 1 for both GNN networks.
We use Adam optimizer~\cite{KingmaB14} with the default parameter setting. In experiments, the batch size is set to 32, the learning rate is 0.001, gradient clipping restricts the gradients within [0,0.1], and the normalization constant $Z_{v,r}$ of R-GCN in Eq.~\ref{r-gcn} is 1. During pre-training, we directly optimize the MIM loss as Section~\ref{MIM}. While, during fine-tuning, the weight $\lambda$ of the MIM loss in Eq.~\ref{CE-rec} is 0.025. 
Our code is publicly available via the link: \textcolor{blue}{\url{https://github.com/RUCAIBox/KGSF}}.

\subsection{Evaluation on Recommendation Task}
In this subsection, we conduct a series of experiments on the effectiveness of the proposed model for the recommendation task. Table~\ref{rec-result} presents the performance of different methods in the two settings.

\subsubsection{Evaluation on All Data Setting}
We first consider the \emph{all data} setting. As we can see, Popularity achieves a comparable performance with TextCNN. 
For our CRS task, utterance text is likely to be sparse and noisy, while Popularity utilizes global statistics for recommending popular items.
Second, the two CRS models Redial and KBRD perform better than Popularity and TextCNN. 
Compared with TextCNN, Redial and KBRD only utilize the entities or items in context to make recommendations. 
Furthermore, KBRD performs better than Redial, since it incorporates external KG information. 
It indicates that KG data is useful to enhance the data representations and improves the performance of the CRS task.
Finally, our model KGSF outperforms the baselines with a large margin. KGSF has incorporated both word-oriented and entity-oriented KGs, and further fuses the two KGs for enhancing the data representations.
% The major reason is that KGSF has considered both kinds of data signals (\ie text and item), aligns the two semantic spaces and mutually enhances their data representations. By utilizing the enhanced representations, our model is more capable of learning user preference.

\subsubsection{Evaluation on Cold Start Setting}
For the \emph{cold start} setting, first, the heuristic method Popularity performs very well, even better than TextCNN and ReDial in most cases.
A possible reason is that in real-world recommender systems, 
a new user is likely to adopt a popular item. 
When there are no items or attributes mentioned in the context, the performance of KBRD and Redial has decreased substantially. 
As a comparison, our model KGSF still performs stably and achieves the best performance among all the methods.
The major reason is that it not only utilizes item-oriented KG, but also uses word-oriented KG. By aligning the two semantic spaces, it can capture important evidence from utterance text to infer user preference in the cold start setting.

\subsubsection{Ablation Study}
In our model, we have incorporated two external KGs and adopted the Mutual Information Maximization method for KG fusion. Here, we would like to examine the contribution of each part.
We incorporate two variants of our model for ablation analysis, namely \emph{KGSF w/o MIM} and \emph{KGSF w/o DB}, which remove the MIM loss and the DBpedia KG, respectively.
Overall, we can see that both components contribute to the final performance. 
Besides, after removing the item-oriented KG, the performance decreases more significantly. It is because once it was removed, the corresponding fusion component with MIM is also removed.

\ignore{
\begin{figure}
\center
\subfigure[Recall@10 in test set]{
\includegraphics[width=0.215\textwidth]{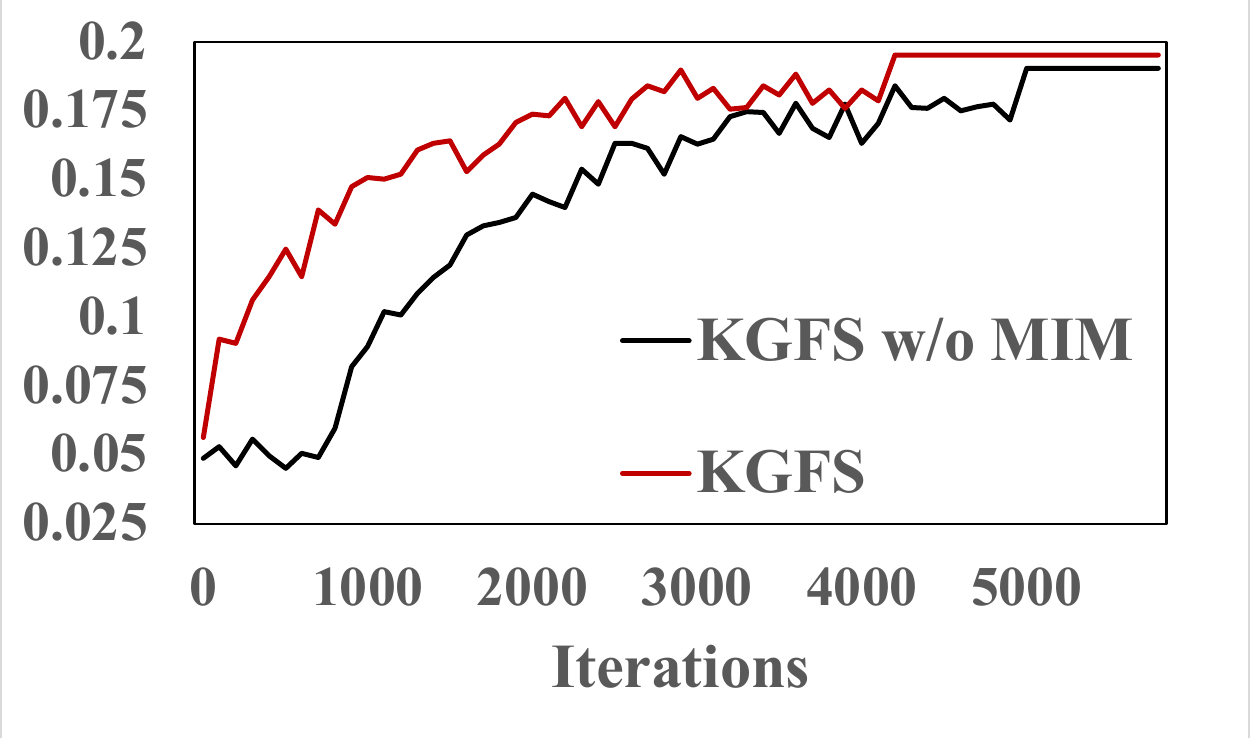}
\label{Recall@10}
}
\subfigure[Recall@50 in test set]{
\includegraphics[width=0.215\textwidth]{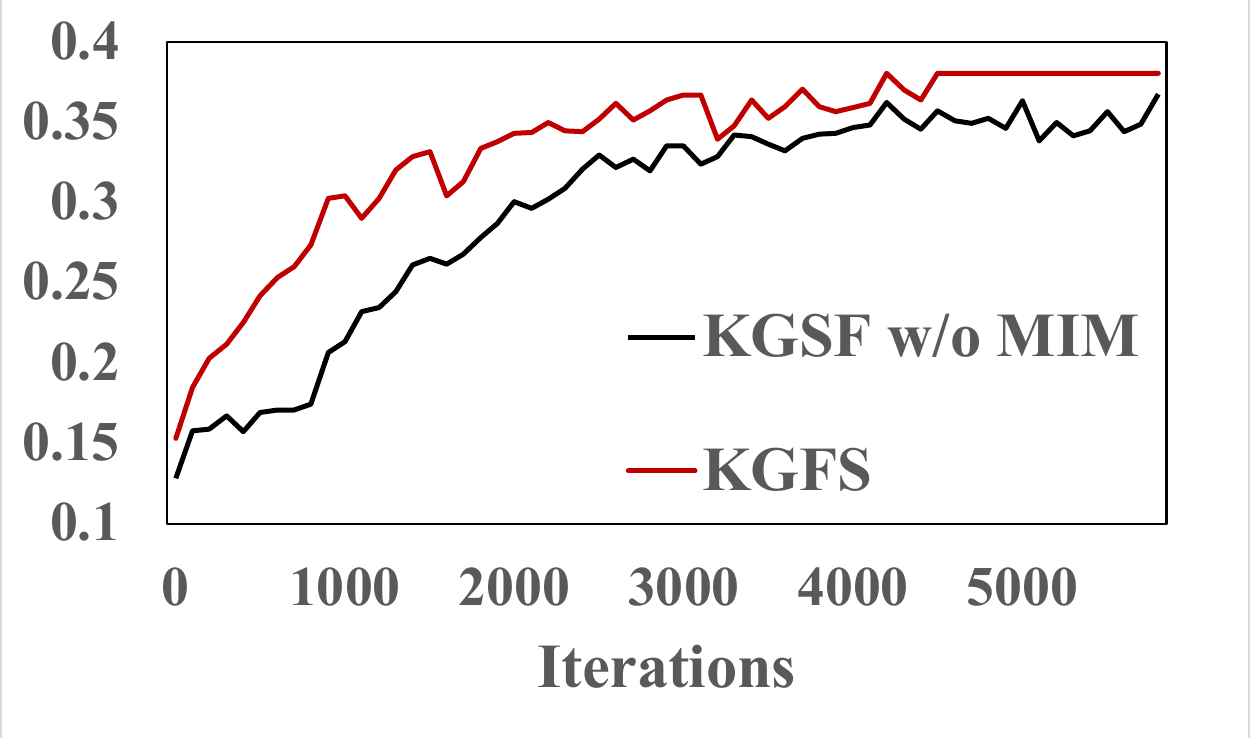}
\label{Recall@50}
}
\caption{Recall@10 and Recall@50 of KGFS \emph{with} and \emph{without} the MIM loss on test set.}
\label{parameter}
\end{figure}
}

\begin{figure}
\center
\subfigure[Recall@10 in test set]{
\includegraphics[width=0.3\textwidth]{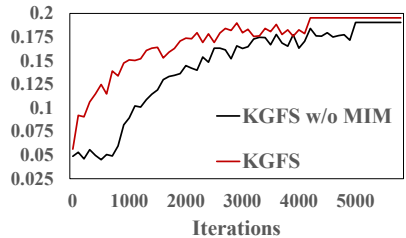}
\label{Recall@10}
}
\subfigure[Recall@50 in test set]{
\includegraphics[width=0.3\textwidth]{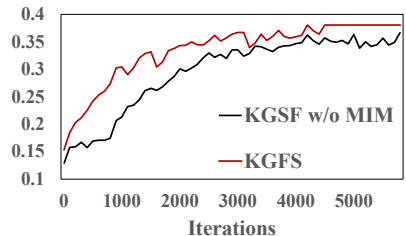}
\label{Recall@50}
}
\caption{Performance (Recall@10 and Recall@50) comparison of KGFS \emph{with} and \emph{without} the MIM loss on test set.}
\label{parameter}
\end{figure}

\subsubsection{The Effect of MIM Technique} 
We adopt the MIM technique to fuse semantic representations from two KGs. As shown in previous experiments, it is useful to improve the performance of our model on the recommendation task. Here, we would like to study whether its improvement is consistent and stable with the increase of the iteration number. We gradually increase the number of iterations for our model on the training set, and report the corresponding performance on the test set. Figure~\ref{parameter} shows how the performance of our model varies with the increase of iterations. We can see that with the MIM technique, our model can achieve a good result with fewer iterations compared to the variant without MIM. Overall, besides the performance improvement, the MIM technique is useful to improve the stability of the training process.

\begin{table}
\caption{Automatic evaluation results on the conversation task. We abbreviate Distinct-2,3,4 as Dist-2,3,4. Numbers marked with * indicate that the improvement is statistically significant compared with the best baseline (t-test with p-value $< 0.05$).}\label{gen-result}
%\small
\centering
\begin{tabular}{lccccc}
     Models &Dist-2 &Dist-3 &Dist-4 &Item Ratio \\
    \hline
    \texttt{Transformer} & 0.148 &0.151 &0.137 &0.194 \\
    \texttt{ReDial} & 0.225& 0.236& 0.228& 0.158 \\
    \texttt{KBRD} & 0.263& 0.368& 0.423 & 0.296 \\
    \hline
    \textbf{\texttt{KGSF}} & \textbf{0.289}*& \textbf{0.434}*& \textbf{0.519}*& \textbf{0.325}* \\
    %\hline
    %\texttt{KGSF w/o KG-L} & 0.150 & 0.156 &0.147 &0.173 \\
    %\texttt{KGSF w/o copy} & 0.249 & 0.382 &0.487 &0.212 \\
    %\texttt{KGSF w/o MIM} &0.232 &0.337 &0.423 &0.289 \\
    \hline
  \end{tabular}
\end{table}

\subsection{Evaluation on Conversation Task}
In this subsection, we construct a series of experiments on the effectiveness of the proposed model on the conversation task. 

\subsubsection{Automatic Evaluation}
We present the results of the automatic evaluation for different methods in Table~\ref{gen-result}.
First, ReDial performs better than Transformer in Distinct-2/3/4, since it utilizes a pre-training RNN model to encode history utterances. While, Transformer performs better with the metric of \emph{Item Ratio}. A possible reason is that Transformer architecture adopts the self-attention mechanism for capturing temporal pairwise interaction, which is more suitable to model the relations between words and items than RNN and CNN.
Second, among the three baselines, KBRD generates the most diverse responses and achieves the highest item ratio, \ie containing more mentions of items in the generated text.
This model utilizes KG information to promote the predictive probability of entities and items.
Compared with these baselines, our KGSF model is consistently better in all evaluation metrics.
KGSF has utilized the KG information in two major steps, namely the knowledge-enhanced Transformer decoder and the copy mechanism, which enhances the informativeness of the generated text.

\begin{table}
%\small
\caption{Human evaluation results on the conversation task. Numbers marked with * indicate that the improvement is statistically significant compared with the best baseline (t-test with p-value $< 0.05$). }\label{gen-result-human}
\centering
\begin{tabular}{lccc}
     Models &Fluency &Informativeness\\
    \hline
    \texttt{Transformer} &0.92 &1.08\\
    \texttt{ReDial} &1.37 &0.97\\
    \texttt{KBRD}  &1.18 &1.18\\
    \hline
    \textbf{\texttt{KGSF}} & \textbf{1.54}*& \textbf{1.40}* \\
    \hline
  \end{tabular}
\end{table}

\begin{figure}
\includegraphics[width=0.45\textwidth]{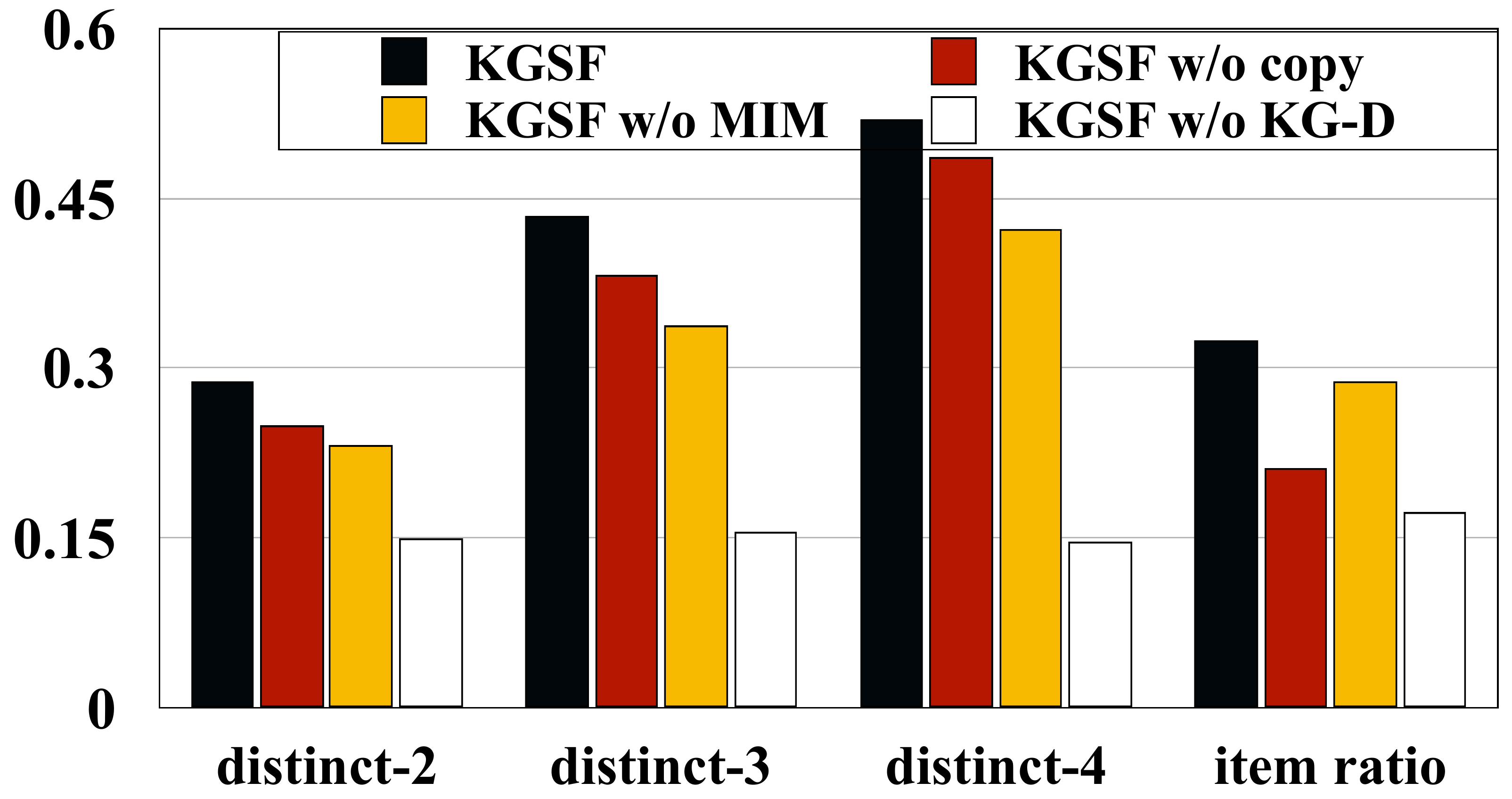}
\caption{Ablation study on conversation task.}
\label{ablation}
\end{figure}

\subsubsection{Human Evaluation}
Table~\ref{gen-result-human} presents the result of human evaluation for the conversation task.
First, among the three baselines, ReDial performs best in terms of the metric of \emph{Fluency}, since it utilizes a pre-training encoder on multiple language tasks~\cite{DBLP:conf/iclr/SubramanianTBP18}. However, we find that it tends to generate short and repetitive responses. This is the so-called ``safe response" issue in the dialog generation task. Without additional supervision signal, it is likely to overfit to the frequent utterances in the training set. 
Second, KBRD performs best in terms of \emph{Informativeness} score among the three baselines. It utilizes KG data to promote the probability of low-frequency words.
Finally, our proposed model KGSF is consistently better than all the baselines with a large margin. We carefully design a KG-enhanced Transformer decoder. Our model is able to utilize contextual information effectively, and generate fluent and informative responses.

\subsubsection{Ablation Study}
We also conduct the ablation study based on three variants of our complete model, include: (1) \emph{KGSF w/o KG-D} by removing the KG-based attention layers from the Transformer decoder, (2)  \emph{KGSF w/o copy} by removing the copy mechanism, and (3) \emph{KGSF w/o MIM} by removing the MIM loss.
As shown in Figure~\ref{ablation}, first, all the techniques are useful to improve the final performance. Besides, the KG-based attention layer seems to be more important in our task, yielding a significant decrease when removed.  KG-based attention layers can effectively inject the fused KG information into the decoder by multi-head attention mechanism.
For the MIM loss, besides its contribution to the recommendation task  (See Table~\ref{rec-result}), it also improves the quality of the generated responses, which indicates its usefulness for KG-based semantic fusion. 

\begin{figure}
	\includegraphics[width=0.45\textwidth]{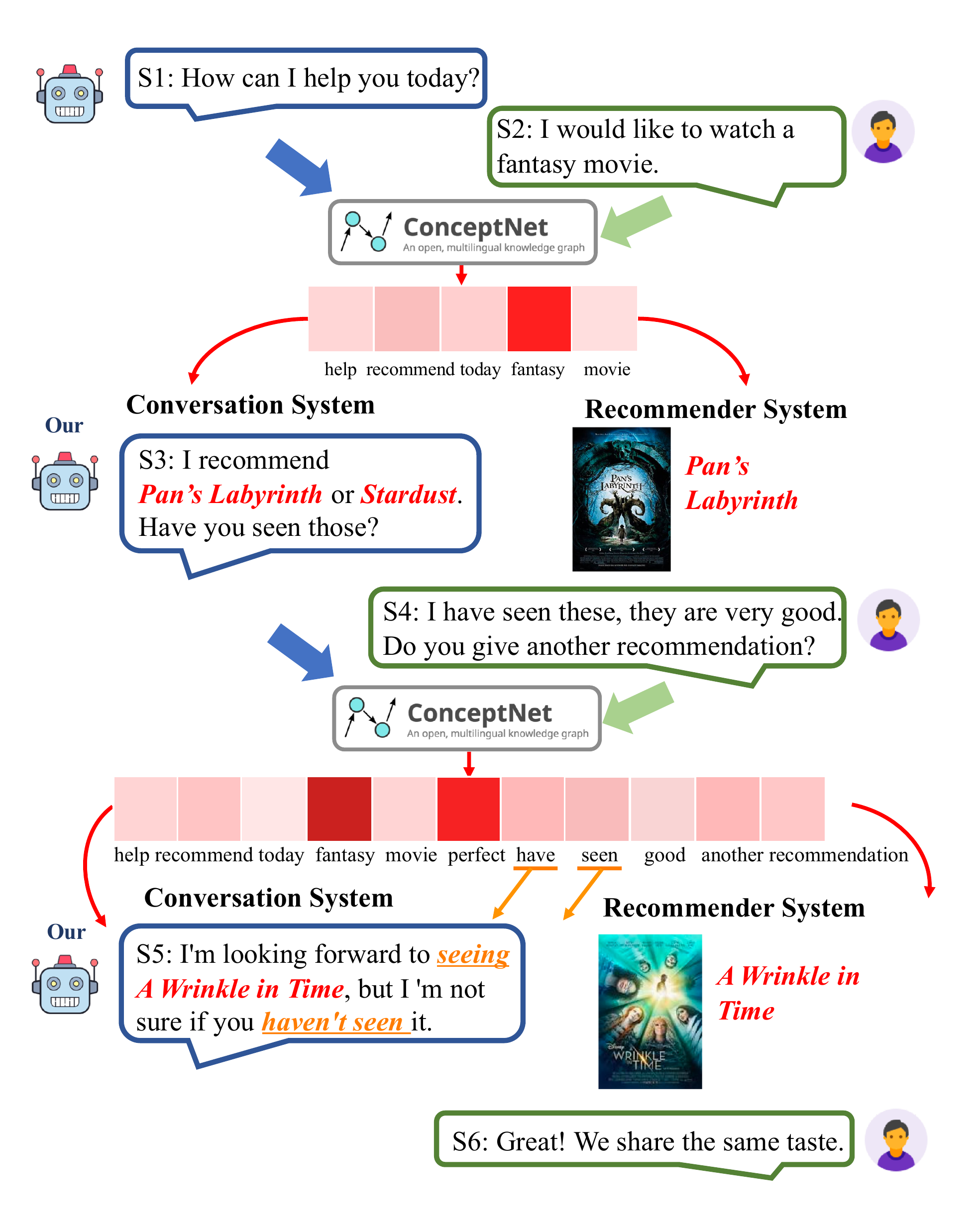}
	\caption{A sampled conversation with six-turn utterances between our CRS agent (recommender) and a real user (seeker). We use color bars to indicate attention weights of words in the recommendation component. The first-round recommendation is unsuccessful, while the second-round recommendation is successful. }
	\label{casestudy}
\end{figure}

\subsection{Qualitative Analysis}
In this part, we present a qualitative example to illustrate how our model works in practice.

In Fig.~\ref{casestudy}, a user requests the recommendations on \emph{fantasy movies}, and our system accurately identifies the key word ``\emph{fantasy}" by assigning a larger attention weight.
The attention weights are computed by the self-attentive mechanism in Eq.~\ref{eq-SA} based on the KG-enhanced word embeddings.
With the focused preference of \emph{fantasy}, our recommender component returns the candidate \emph{``Pan's Labyrinth''}. While, interestingly, our dialog component not only includes the mentions of the recommended movie, but also generates another related movie (\emph{``Stardust''}) in the utterance. Receiving the response, the user rejects the recommendation since she/he has watched both movies before.
Then, our recommender component updates the representation of user preference, and returns another recommendation. Although the words of ``\emph{have}" and ``\emph{seen}" received a small attention weight by the recommender component, our dialog component also boosts their weights since they are helpful to generate a more informative reply.
\section{Conclusion and Future Work}
In this paper, we proposed a novel KG-based semantic fusion approach for CRS. By utilizing two external KGs, we enhanced the semantic representations of words and items, and used Mutual Information Maximization to align the semantic spaces for the two different components. 
Based on the aligned semantic representations, we developed a KG-enhanced recommendation component for making accurate recommendations, and a KG-enhanced dialog component that can generate informative keywords or entities in the utterance text. 
By constructing extensive experiments, our approach yielded better performance than several competitive baselines.

As future work, we will consider using more kinds of external information to improve the performance of CRS, \eg user demographics~\cite{zhao-kdd-2014}. 
Besides, we will investigate how to make the utterance more persuasive and explainable for the recommendation results. Finally, another interesting topic is how to incorporate historical user-item interaction data and start the conversation with a pre-learned user profile. 

\section*{Acknowledgement}
This work was partially supported by the National Natural Science Foundation of China under Grant No. 61872369 and 61832017,  Beijing Academy of Artificial Intelligence (BAAI) under Grant No. BAAI2020ZJ0301, and Beijing Outstanding Young Scientist Program under Grant No. BJJWZYJH012019100020098, the Fundamental Research Funds for the Central Universities, the Research Funds of Renmin University of China under Grant No.18XNLG22 and 19XNQ047. Xin Zhao is the corresponding author.

%\bibliographystyle{ACM-Reference-Format}
%\bibliography{sample-base}
%%% -*-BibTeX-*-
%%% Do NOT edit. File created by BibTeX with style
%%% ACM-Reference-Format-Journals [18-Jan-2012].

\end{document}